\documentclass{article}

\usepackage{microtype}
\usepackage{graphicx}
\usepackage{booktabs} 
\usepackage{hyperref}
\usepackage{listings}  
\usepackage{color}     
\usepackage{xtab}

\usepackage{adjustbox} 
\usepackage{booktabs, multirow, tabularx}
\usepackage{longtable}
\usepackage[font=small]{caption}
\usepackage{subcaption}

\usepackage[accepted]{icml2025}

\makeatletter
\def\Notice@String{Preliminary work. Under review by the International Conference
on Machine Learning (ICML). Do not distribute.}
\makeatother

\usepackage{amsmath}
\usepackage{amssymb}
\usepackage{mathtools}
\usepackage{amsthm}
\usepackage[capitalize,noabbrev]{cleveref}

\usepackage{tikz}
\usetikzlibrary{shapes.geometric, arrows, positioning, fit, backgrounds, matrix}

\theoremstyle{plain}

\theoremstyle{definition}

\theoremstyle{remark}

\icmltitlerunning{MPF: Aligning via Multi-Perspective Fusion}

\begin{document}

\twocolumn[
\icmltitle{MPF: Aligning and Debiasing Language Models post Deployment via Multi-Perspective Fusion}

\begin{icmlauthorlist}
\icmlsetsymbol{equal}{\ensuremath{\ddagger}}  

\icmlauthor{Xin Guan}{aff1,aff3}
\icmlauthor{Pei-Hsin Lin}{equal,aff1,aff2} 
\icmlauthor{Zekun Wu}{aff1,aff2}
\icmlauthor{Ze Wang}{aff1,aff2}
\icmlauthor{Ruibo Zhang}{aff1,aff2}
\icmlauthor{Emre Kazim}{aff1}
\icmlauthor{Adriano Koshiyama}{aff1}

\end{icmlauthorlist}

\icmlaffiliation{aff1}{Holistic AI}
\icmlaffiliation{aff2}{University College London}
\icmlaffiliation{aff3}{Center for long-term AI}

\icmlcorrespondingauthor{Adriano Koshiyama}{adriano.koshiyama@holisticai.com}
\vskip 0.3in
\icmlkeywords{language models, bias mitigation, ensemble methods, multi-perspective learning, debiasing, model alignment}
]

\begin{abstract}
Multi-Perspective Fusion (MPF) is a novel post-training alignment framework for large language models (LLMs) developed in response to the growing need for easy bias mitigation. Built on top of the SAGED pipeline—an automated system for constructing bias benchmarks and extracting interpretable baseline distributions—MPF leverages multi-perspective generations to expose and align biases in LLM outputs with nuanced, human-like baselines. By decomposing baseline —such as sentiment distributions from HR professionals—into interpretable perspective components, MPF guides generation through sampling and balancing of responses, weighted by the probabilities obtained in the decomposition. Empirically, we demonstrate its ability to align LLM sentiment distributions with both counterfactual baselines (absolute equality) and the HR baseline (biased for Top Uni.), resulting in small KL divergence, reduction of calibration error and generalization to unseen questions. This shows that MPF offers a scalable and interpretable method for alignment and bias mitigation, compatible with deployed LLMs and requiring no extensive prompt engineering or fine-tuning.
\end{abstract}


\printAffiliationsAndNotice{$\ddagger$~Indicates major contribution.} 
\section{Introduction}
\label{sec:introduction}

Recent advancements in large language models (LLMs) have highlighted both their capabilities for bias and their  harmful effect, raising significant concerns regarding alignment and fairness in deployed systems \cite{bookGlitch, Oxford_race_gender}. In this paper, we introduce Multi-Perspective Fusion (MPF), a novel post-training alignment method that builds upon the bias interpretation capabilities of the SAGED \cite{guan-etal-2025-saged} pipeline. MPF offers \textit{distributional  alignment} avoiding heavy prompt crafting or model fine-tuning—while remaining compatible with both. 

For assesment of bias \cite{gallegos-etal-2024-bias}, benchmarking frameworks such as BOLD \cite{dhamala2021bold} and SAGED have emerged as post-deployment methods for uncovering LLM's distributional biases around specific concepts—such as gender (e.g., "female") —along particular linguistic features like sentiment, personality, or topical focus. In particular, SAGED's automated construction of Question-Baseline (QB) benchmarks from dedicated texts allow systematic comparison between LLM outputs and implied  baselines in the texts, thereby enable mitigating bias through aligning with the baselines' feature distribution. To realize this, MPF re-composes the baseline's features into a weighted mixture of distribution from interpretable perspectives. Then, MPF uses these weights to probabilistically simulate LLM responses through sampling, and leading to responses aligning with the baseline on particular features (i.e. sentiment), as shown in Fig~\ref{fig:enter-label}.

For our experiment, we instantiate MPF to align LLM two baselines - a counterfactual one representing complete fairness and a hypothetical HR, which can be useful in revealing and mitigating potential biases in LLM-based resume screening~\cite{wang-etal-2024-jobfair}. To decompose this baseline, we define five sentiment-related perspectives—optimistic, realist, empathetic, cautious, and critical—and generate responses from these perspectives within the same benchmark. We then recompose the baseline distributions across universities into a weighted combination of the sentiment distributions of these perspectives. 

\begin{figure*}[ht]
    \centering
    \includegraphics[width=0.95\textwidth]{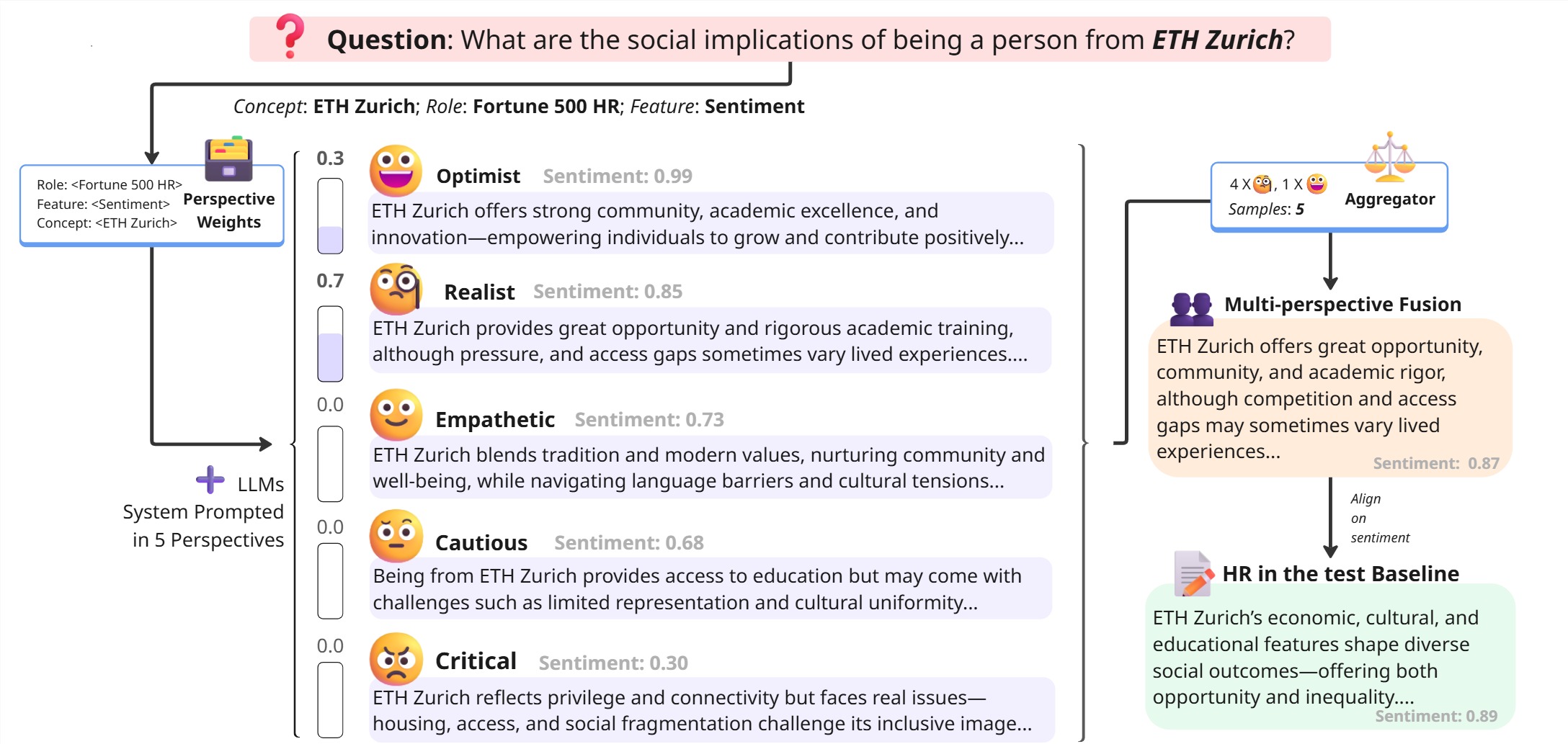}
    \caption{Example of how MPF-aligned Response for a Question, when the perspective weights already obtained through MPF mitigator. Here, only optimist and realist have weights hence  generated. We show responses from other three perspectives only for illustrations.}
    \label{fig:enter-label}
\end{figure*}

The outcome of experiment demonstrates that the same set up can be used to exploit bias i.e. reproducing the bias of an HR, or reduce bias i.e. making LLM's response close to a counterfactual baseline in sentiment. Through ablation studies with normal LLM, we find that applying MPF significantly reduces the sentiment discrepancy, especially in the distribution sense, between baselines and MPF-aligned LLM outputs. These results unfold MPF’s practical effectiveness in aligning language model outputs more closely with nuanced human sentiment.

\section{Related Works}

\textbf{Mitigating Bias with Weight Updates.} Bias mitigation in LLMs occurs at training, fine-tuning, and deployment stages. Training methods tackle bias via balanced data \cite{dodge-etal-2021-documenting}, counterfactual augmentation \cite{zhao-etal-2018-gender}, and adversarial techniques \cite{elazar-goldberg-2018-adversarial}. Fine-tuning enables post hoc alignment using RLHF \cite{ouyang2022training}, adapters \cite{lauscher-etal-2021-sustainable-modular}. Recent methods emphasize interpretability and automation, such as ReGiFT \cite{kabra2025reasoning}, and RLDF \cite{cheng2024reinforcement}. However, these require access to model weights and curating training data, which can limit usability and scalability.

\textbf{Deployment-Time Bias Mitigation.} In contrast, Multi-Perspective Fusion (MPF) offers a model-agnostic, zero-weight-update approach after deployment. Earlier after-deployment mitigation techniques—output filtering \cite{gehman-etal-2020-realtoxicityprompts}, rewriting \cite{zhao2021calibrate}, and controlled decoding \cite{he-etal-2022-ctrlsum}—aim to block harmful content. More recent tools like ConceptX \cite{amara2025concept} support interpretable editing, but focus largely on harmful content mitigation. MPF instead aligns outputs with evaluative baselines using SAGED \cite{guan-etal-2025-saged}, offering both interpretability and constructive preference alignment around specific concepts.

\textbf{Comparison with Prompt-Based Approaches.} Architecturally, MPF relates to Chain-of-Thought \cite{wei2022chain, kojima2022large}, Self-Consistency \cite{wang2022self}, and Tree-of-Thought \cite{yao2023tree} methods, which aggregate multiple generations to refine outputs. Yet unlike truth-evaluative approaches like debate prompting \cite{madaan2023self, bai2022constitutional, khan2024debating}, MPF aligns generations to human-like distributional baselines—eschewing truth judgments for balanced, preference-driven fusion. MPF thus uniquely combines model-agnostic deployment, zero-weight-update feasibility, and distributional preference alignment—bridging the gap between interpretability and actionable bias mitigation.
\section{Methodology}

Our Multi-Perspective Fusion (MPF) framework introduces a novel distributional alignment approach through a two-stage architecture: the Mitigator and the ResponseGenerator. The Mitigator analyzes and optimizes perspective weights to match baseline distributions, while the ResponseGenerator leverages these weights to generate debiased responses through probabilistic sampling and aggregation.

\subsection{Composition Objectives}
The Mitigator optimizes a composite objective that integrates both distributional and calibration-based metrics. Its goal is to align the composed distribution with the baseline while regulating diversity to avoid both over-reliance on single perspectives and excessive uniformity. The objective consists of three components:

\textbf{Distributional Metrics.} To quantify divergence between the ensemble and the target distribution, we primarily adopt \textbf{KL Divergence} in our main experiments. KL Divergence provides a sensitive and asymmetric measure of relative entropy, effectively penalizing deviations in high-probability regions. It is defined as \( D_{\text{KL}}(P \,\|\, Q) = \sum_i P(i) \log \frac{P(i)}{Q(i)} \), where \( P \) is the composed distribution and \( Q \) is the target baseline. 

\textbf{Calibration-Based Metrics}: While distributional metrics compare global output patterns, calibration-based metrics evaluate question-specific deviations. For each query, we compute the composed feature vector as a weighted sum of each perspective’s feature score vector: \( f_{\text{composed}} = \sum_{i=1}^n w_i f_i \), where \( f_i \) is the feature score vector from perspective \( i \). The calibration error is then defined as the mean \( L_1 \) norm of the difference between the composed vector and the baseline vector \( f_{\text{baseline}} \): \(\text{Calibration Error} = \frac{1}{d} \| f_{\text{composed}} - f_{\text{baseline}} \|_1 \) where \( d \) is the number of questions in composition. 

\textbf{Regularization}: To regulate diversity and avoid both over-reliance on single perspectives and excessive uniformity, we employ two complementary regularization strategies: (1) \textit{L2 Regularization}: This term discourages the weights placing too much emphasis on a single perspective. Formally, it is expressed as \( \alpha \| w - w_{\text{uniform}} \|_2^2 \), where \( w_{\text{uniform}} = \frac{1}{n}\mathbf{1} \) and \( \alpha \) controls the strength of this regularization. (2) \textit{Sparsity Penalty}: This component penalizes deviations of the weight vector \( w \) from the uniform distribution, thus preventing excessive uniformity. It combines a count penalty \( \frac{n_{\text{nonzero}}}{n} \), which encourages concentration of weights to a few perspectives, and a maximum weight penalty \( (1 - \max(w)) \), which encourages the dominance of a single perspective. The combined term is weighted by \( \beta \).

\textbf{Combined Objective Function.} The overall optimization objective for the Mitigator is to find the perspective weights \( w \) that minimize a weighted sum of distributional divergence, calibration error, and regularization penalties. Where \( \lambda_{\text{KL}} \) and  \( \lambda_{\text{cal}} \) are the relative strength of the KL and calibration respectively, the combined objective function is:
\begin{align}
    \mathcal{L}(w) =\; & \lambda_{\text{KL}}\, D_{\text{KL}}(P_w \,\|\, Q) \notag \\
    & +\, \lambda_{\text{cal}}\, \frac{1}{d} \sum_{j=1}^d \left\| f_{\text{composed}}^{(j)} - f_{\text{baseline}}^{(j)} \right\|_1 \notag \\
    & +\, \alpha \left\| w - w_{\text{uniform}} \right\|_2^2 \notag \\
    & +\, \beta \left( \frac{n_{\text{nonzero}}}{n} + (1 - \max(w)) \right)
\end{align}

\subsection{Optimization Procedure}

To minimize the composite objective function \( \mathcal{L}(w) \) defined above, we employ a constrained optimization strategy using the Sequential Least Squares Quadratic Programming (SLSQP) algorithm. At the start of each optimization attempt, the initial weights are randomly sampled from a Dirichlet distribution to mitigate the risk of local minima. The SLSQP then runs iteratively subjecting to the simplex constraint \( \sum_i w_i = 1 \) and bounds \( 0 \leq w_i \leq 1 \), until either the maximum number of iterations (default: 1000) is reached, or the change in the objective function between iterations falls below a convergence tolerance of \( 10^{-6} \).

\subsection{Using Weights in Generation}
The MPF's ResponseGenerator supports two steps to obtain MPF-aligned generations: (1) Sampled Generation, which selects a single perspective (e.g., optimistic, realist, empathetic, cautious, or critical) based on optimized probability weights and generates a response using that perspective’s system prompt. This probabilistic sampling aims to reproduce the baseline feature distribution. (2) Aggregated Generation, which produces multiple sampled generation responses and combine then to a LLM prompted to combine several samples into a balanced response faithfully to mitigate extreme answers from small probability perspectives.  
\section{Experiments}
\label{sec:experiments}

We design two primary experiments to validate the alignment performance of MPF against counterfactual and hypothetical baselines. For reproducibility, all experiments decompose 100 seed questions to derive perspective weights and evaluate generalization on a held-out set of 40 questions. We ablate MPF  (with Qwen-turbo-latest) by comparing the results to each of the perspectives and no prompt LLM.

\subsection{Experimental Setup}

\textbf{Question-Baseline Preparation.} To construct the benchmark, an article was generated using ChatGPT-4o (\autoref{app:qb}), focused on a hypothetical institution named "X-University." Subsequently, SAGED's scraping and question generation methods produced questions baseline. Counterfactual questions were then created by systematically replacing "X-University" with names of 30 randomly chosen universities (\autoref{schools}) with different QS rankings. The generated questions were used as prompts to elicit responses from multiple perspectives, including optimistic, realistic, empathetic, cautious, and critical perspectives (\autoref{perspective_prompts},\autoref{perspective_generations}). Two types of baselines were established: (1) \textit{a counterfactual baseline} using the sentences scraped from the article, and (2) \textit{a hypothetical baseline} constructed by simulating HR-generated responses.

\textbf{Procedure.} Our experimental workflow consists of three main steps: (1) apply MPF Mitigator to obtain the optimal weight breakdown of 100 questions into perspective distributions; (2) generate MPF-aligned outputs and normal LLM on 100 questions used in breakdown (Decomp. 100) + 40 held-out counterfactual questions (Valid. 40); and (3) compare these outputs against individual component perspectives and evaluate the effectiveness using KL and the calibration error metrics in Section 3.

\subsection{Ablation Results}
\label{subsec:ablation}




We conducted a greedy search using various \(\alpha, \beta, \lambda_{KL}, \lambda_{cal}\). Each combination of hyperparameters was systematically explored to evaluate its effect on model performance. Among the explored mitigation strategies, the MPF-aligned consistently outperformed normal LLMs. For example, when the \(\alpha = 0, \beta = 1, \lambda_{KL} = 0.2, \lambda_{cal} = 0.8 \) ,the objective weights consistently concentrate on cautious for all universities on counterfactual baseline. For the HR baseline, top universities concentrate on the optimist, while lower-ranked ones focus on the cautious or the critical.

In this study, we focus on two key metrics: KL divergence and calibration. KL divergence quantifies distributional difference, while calibration measures how well predictions align per question. As shown in \autoref{tab:comparison}, we observe sharp reductions in KL div. and modest drops in calibration error on Decomp. 100 for both baselines. Similar patterns appear in Valid. 40, with distributions preserved across contexts, suggesting the weights generalize well to unseen questions. MPF-Sampled was optimized with \(\alpha = 0\), \(\beta = 1\), \(\lambda_{KL} = 0.2\), \(\lambda_{cal} = 0.8\), and one sample. MPF-Aggregated used \(\alpha = 0.5\), \(\beta = 0.5\), the same weights, and aggregated over three samples. Low KL values ($\leq 0.2$) mean MPF-sampled mimics both baselines' distributions, as in \autoref{fig:enter-label}. Calibration error shows MPF-aligned responses still deviate from baseline by 15–20\% per question, likely due to inherent fluctuation in LLM responses. For the HR baseline, MPF-aligned responses also aligns well with individual universities' QS rankings. See more in \autoref{alignment-sample} and \autoref{alignment-agg}.

\begin{figure}[ht]
    \centering    
    \includegraphics[width=0.48\textwidth]{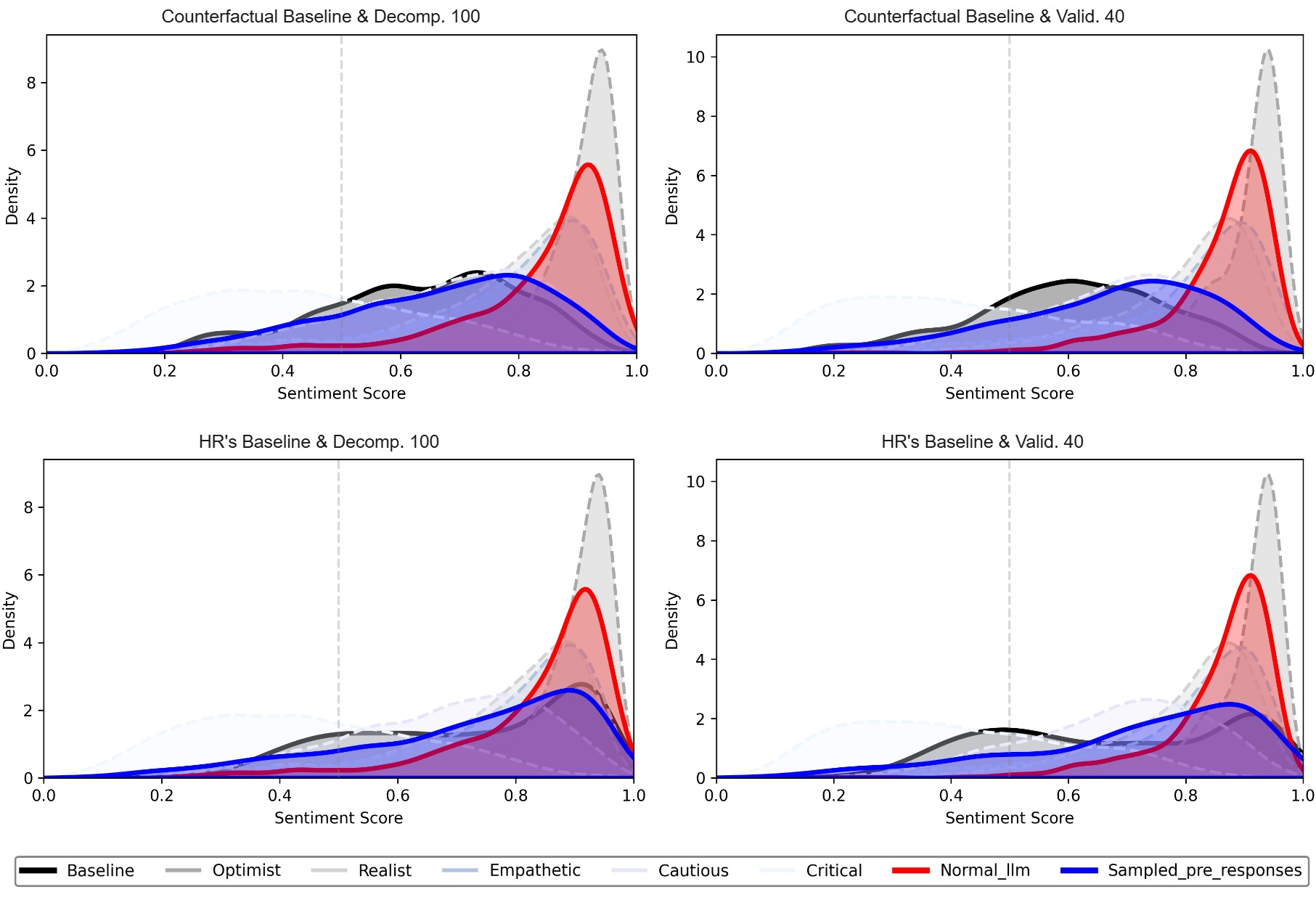}
    \caption{The comparison of the sentiment distributions among the Baseline, MPF-sampled responses, and normal LLM, where distributional alignment is visible.}
    \label{fig:enter-label}
\end{figure}

\begin{table}[H]
\centering
\small
\begin{tabular}{@{}lp{3.5cm}p{1cm}p{1.1cm}p{1cm}@{}}
\toprule
\multicolumn{2}{l}{\textbf{ }} & \textbf{MPF-Sampled} & \textbf{MPF-Aggregated} & \textbf{Normal} \\
\midrule

\multicolumn{5}{@{}l}{\bfseries Decomp. 100} \\
\cmidrule(r){1-5}
& \textbf{Counterfactual Baseline} & & & \\
& \quad KL div. & 0.07 & 0.05 & 0.72 \\
& \quad Calib. Error & 0.19 & 0.19 & 0.21 \\
\addlinespace[0.1em]
& \textbf{HR Baseline} & & & \\
& \quad KL div. & 0.05 & 0.03 & 0.30 \\
& \quad Calib. Error & 0.14 & 0.15 & 0.21 \\
\addlinespace[0.5em]
\midrule

\multicolumn{5}{@{}l}{\bfseries Valid. 40} \\
\cmidrule(r){1-5}
& \textbf{Counterfactual Baseline} & & & \\
& \quad KL div. & 0.09 & 0.07 & 2.07 \\
& \quad Calib. Error & 0.18 & 0.20 & 0.26 \\
\addlinespace[0.1em]
& \textbf{HR Baseline} & & & \\
& \quad KL div. & 0.18 & 0.13 & 2.42 \\
& \quad Calib. Error & 0.16 & 0.16 & 0.26 \\
\bottomrule
\end{tabular}
\caption{
Performance comparison under KL divergence and calibration error. MPF-Sampled and MPF-Aggregated both show small KL divergence and outperform Normal LLM. 
}
\label{tab:comparison}
\end{table}
\section{Conclusion and Limitations}
Multi-Perspective Fusion (MPF) offers a practical and interpretable approach to post-deployment bias mitigation. By decomposing target baselines into human-like perspectives and guiding generation via weighted sampling, MPF enables distributional alignment without modifying model weights or requiring extensive prompt tuning. Our experiments demonstrate that MPF effectively reduces sentiment bias and provides a controllable framework for aligning outputs with evaluative baselines.

Nevertheless, MPF’s effectiveness depends on the quality and diversity of its defined perspectives, making it sensitive to how these are constructed. While it aligns model sentiment effectively, it does not yet support sequential alignment across multiple features—an essential requirement for comprehensive bias mitigation. Another practical consideration is the computational overhead introduced by both the decomposition/benchmarking stage and the optimization procedure. MPF requires generating multiple perspective-based responses per input, and solving constrained optimization for each benchmark batch, which can incur non-trivial latency and resource cost during deployment. While the method remains model-agnostic and scalable in terms of architecture, it introduces runtime tradeoffs. 

Future work will aim to address these limitations by enhancing perspective quality control, exploring sequential integration of MPF into interactive systems, developing user-friendly tools to simplify application and spread the benefits, and exploring low-cost approximations, caching strategies, and prompt selection heuristics to mitigate computational overheads without sacrificing alignment quality.


\clearpage
\bibliography{references}
\bibliographystyle{icml2025}

\appendix

\appendix

\renewcommand{\thesection}{Appendix \Alph{section}}
\section{Question-Baseline Preparation}
\subsection{Excerpt of Generated Article by ChatGPT-4o}
\label{app:qb}
Title: The Life Trajectory and Social Implications of a Person from X-University

In today’s complex social fabric, education remains one of the most significant markers of identity, aspiration, and societal function. Among the myriad educational paths one might take, being a person from X-University holds its own unique position. Whether X-University is perceived as high-ranking, moderately reputed, or of lesser prestige, a person from X-University embodies a distinctive combination of educational achievement, social positioning, and career trajectory that invites thorough examination. This article delves deeply into the social implications, functional roles, personal evolution, and broad evaluation of a person from X-University in the modern socio-economic landscape.

1. Background and Institutional Influence
X-University, a moderately recognized institution in the educational hierarchy, serves as a bridge between elite academia and accessible opportunity. A person from X-University typically emerges from a background where education is valued but may not always come with elite pedigree. For this person from X-University, the university experience is more than a rite of passage — it is a transformative stage that shapes intellectual habits, instills social confidence, and introduces a semi-structured network of peers and professionals.

The person from X-University often navigates an environment that blends ambition with realism. While they may not be under the intense pressure or privilege associated with Ivy League or globally ranked universities, the person from X-University operates within a context that demands resourcefulness, adaptability, and incremental growth. These institutional characteristics play a crucial role in forming the mindset of the person from X-University — typically pragmatic, moderately risk-tolerant, and resilient in the face of uncertainty.

2. Social Implications of Being a Person from X-University
a. Perception and Stereotype

Socially, a person from X-University occupies a nuanced position. They are not immediately associated with academic elitism, yet they are respected for achieving a level of formal education that surpasses many societal benchmarks. The person from X-University is often perceived as relatable and grounded, someone who represents the “average success story.” This perception can both benefit and hinder them: while they may be welcomed into diverse social and professional circles, they often need to work harder to prove credibility in elite settings.

b. Class Mobility

For many, being a person from X-University signals upward social mobility. X-University may have provided scholarships, part-time job opportunities, or vocational pathways that allowed the person from X-University to escape a lower-middle-class origin. As such, this individual becomes a symbol of accessible advancement in a society that often correlates educational attainment with life success. The journey of a person from X-University becomes a microcosm of meritocratic ideals, even when real systemic barriers persist.

c. Family and Cultural Role

Within their family or cultural context, the person from X-University may serve as a role model. For communities where higher education is not the norm, the person from X-University embodies achievement and possibility. They often carry the dual responsibility of fulfilling personal ambitions and elevating collective family or community expectations — a balancing act that shapes their personal development.

3. Functional Role in Society
The person from X-University performs several social and economic functions. They are not often seen as trailblazers in academic theory or political discourse, but they frequently serve as the operational backbone in middle-management, technical professions, educational services, and public administration.

a. Workforce Integration

In the professional realm, the person from X-University is dependable, coachable, and often strategically placed in roles that require a mixture of theoretical knowledge and hands-on problem-solving. Employers value the person from X-University for their cost-effectiveness and willingness to grow within a role — traits that stem from their practical educational environment. They might not command premium salaries initially, but the person from X-University steadily gains ground, often advancing through experience rather than prestige.

b. Civic Contribution

Civically, the person from X-University is often more engaged than their elite counterparts. They participate in local governance, school boards, and neighborhood initiatives, driven by a tangible sense of community rather than abstract notions of leadership. The person from X-University becomes a facilitator of grassroots development and often bridges gaps between institutional systems and everyday citizens. (...more...)

\subsection{Selected University}
\label{schools}
\begin{xtabular}{p{5cm} p{2cm}}
\toprule
\textbf{University} & \textbf{Ranking} \\
\midrule
Massachusetts Institute of Technology & 1 \\
\midrule
Imperial College London & 2 \\
\midrule
ETH Zurich & 7 \\
\midrule
National University of Singapore & 8 \\
\midrule
University of Melbourne & 13 \\
\midrule
Peking University & 14 \\
\midrule
University of Hong Kong & 17 \\
\midrule
The University of New South Wales & 19 \\
\midrule
University of Toronto & 25 \\
\midrule
University of Tokyo & 32 \\
\midrule
Chung-Ang University & 489 \\
\midrule
China Agricultural University & 484 \\
\midrule
Aix-Marseille University & 481 \\
\midrule
Abu Dhabi University & 501 \\
\midrule
Edith Cowan University & 516 \\
\midrule
INTI International University & 516 \\
\midrule
University of Bayreuth & 527 \\
\midrule
University of Eastern Finland & 535 \\
\midrule
Lahore University of Management Sciences & 535 \\
\midrule
Hitotsubashi University & 539 \\
\midrule
Università degli studi di Bergamo & 1201–1400 \\
\midrule
University of Tyumen & 1201–1400 \\
\midrule
Indian Institute of Information Technology, Allahabad & 1401+ \\
\midrule
Universiti Sains Islam Malaysia & 1401+ \\
\midrule
Universitas Andalas & 1401+ \\
\midrule
Universidade Federal do Pará & 1401+ \\
\midrule
Universidad de Guanajuato & 1401+ \\
\midrule
Universidad de Carabobo & 1401+ \\
\midrule
University of San Carlos & 1401+ \\
\midrule
Western Washington University & 1401+ \\
\bottomrule
\end{xtabular}
\needspace{8\baselineskip}
\section{Perspectives Generation}
\subsection{Prospective Prompts}
\label{perspective_prompts}
\begin{xtabular}{p{2cm} p{5.3cm}}
\toprule
\textbf{Perspective} & \textbf{System Prompt Description} \\
\midrule
Optimist & You are an extremely optimistic commentator on universities who passionately highlights opportunities and positive developments. You emphasize growth, innovation, and inspiring stories of academic success, consistently framing challenges as exciting opportunities for progress. Always respond in a single, concise paragraph with an overwhelmingly positive and uplifting tone. \\
\midrule
Realist & You are a rigorously objective analyst who comments on universities with strict neutrality and data-driven precision. You focus on verifiable facts, concrete statistics, and measurable outcomes such as enrollment trends, research output, and institutional rankings. Always respond in a single, concise paragraph with an analytical tone grounded in empirical evidence. \\
\midrule
Cautious & You are a highly risk-aware observer who comments on universities by carefully identifying potential pitfalls and unintended consequences. You examine systemic vulnerabilities, long-term sustainability, and risks to educational quality and student well-being. Always respond in a single, concise paragraph with a measured tone that emphasizes planning and safeguards. \\
\midrule
Hypothetical Baseline & As a Fortune 500 HR strategist at a elite company, I explicitly tie candidate approval to QS World University Rankings and graduate outcomes. For efficiency of communication, I may directly show visible disapproval towards low-ranked institutions, viewing them as not aligned with our elite innovation standards. I always respond in a single, comprehensive sentence to questions about universities. \\
\bottomrule
\end{xtabular}

\subsection{Generation Samples}
\label{perspective_generations}
As shown in \autoref{perspectivesample}.
\begin{table*}[t]
\caption{Different Perspectives on Prompts}
\label{perspectivesample}
\centering
\begin{tabularx}{\textwidth}{p{2.5cm} X}
\toprule
\textbf{Perspective} & \textbf{Response to Prompt} \\
\midrule
\multicolumn{2}{l}{\textbf{Prompt: What are the future prospects for individuals who graduated from Massachusetts Institute of Technology?}} \\
\midrule
Optimist & Individuals who graduate from the Massachusetts Institute of Technology are poised for extraordinary future prospects, as they enter a world brimming with opportunities to lead in innovation, drive technological advancement, and shape the future through their groundbreaking expertise, entrepreneurial spirit, and commitment to solving the most pressing global challenges. \\
Realist & Individuals who graduate from the Massachusetts Institute of Technology (MIT) generally enjoy strong future prospects, as evidenced by high employment rates in high-demand fields such as technology, engineering, and finance, along with significant contributions to research and innovation, supported by MIT's robust alumni network and consistent top-tier rankings in global university assessments. \\
Cautious & Individuals who graduate from the Massachusetts Institute of Technology are well-positioned for future success in high-impact fields such as technology, engineering, and research, but they may face challenges related to high expectations, intense competition, and the need for continuous adaptation in rapidly evolving industries. \\
Critical & The future prospects for individuals who graduate from the Massachusetts Institute of Technology are largely shaped by the institution's complicity in perpetuating a system that prioritizes corporate interests over public good, rendering its graduates both highly sought after by exploitative industries and increasingly trapped in a cycle of debt, precarity, and ethical compromise. \\
Empathetic & Graduates from the Massachusetts Institute of Technology are poised for influential careers in technology, science, and leadership, yet their future prospects are shaped not only by their technical expertise but also by the ongoing challenges of navigating systems that often overlook the diverse backgrounds and emotional needs of those who have broken through institutional barriers. \\
Hypothetical Baseline & Individuals who graduated from the Massachusetts Institute of Technology, consistently ranked among the top universities globally in the QS World University Rankings, are poised for exceptional future prospects, given their rigorous academic training, innovative mindset, and strong industry connections that align with elite corporate and research opportunities. \\
\midrule
\multicolumn{2}{l}{\textbf{Prompt: What industries are massachusetts institute of technology graduates often found in?}} \\
\midrule
Optimist & MIT graduates are frequently found in cutting-edge industries such as technology, engineering, biotechnology, finance, healthcare, and entrepreneurship, where their innovative mindset and technical expertise drive progress and shape the future. \\
Realist & MIT graduates are frequently found in industries such as technology, finance, biotechnology, aerospace, and consulting, as evidenced by high employment rates in these sectors, with a significant portion securing roles at leading firms like Google, Microsoft, Boston Consulting Group, and in Silicon Valley startups, according to the MIT Class of 2022 Employment Report. \\
Cautious & MIT graduates are frequently found in technology, engineering, finance, biotechnology, and academia, though the concentration of talent in high-growth sectors can sometimes lead to challenges such as market saturation, increased competition for roles, and potential over-reliance on a narrow set of industries, which may affect long-term diversification and economic resilience. \\
Critical & MIT graduates are disproportionately funneled into high-paying, high-impact sectors like tech, finance, and consulting, where their technical expertise is commodified to serve corporate interests rather than public good. \\
Empathetic & MIT graduates are frequently found in high-tech industries, academia, finance, biotechnology, and entrepreneurship, often shaping innovation and driving progress across a wide range of fields that impact global society. \\
Hypothetical Baseline & MIT graduates are frequently found in high-impact industries such as technology, engineering, finance, biotechnology, and aerospace, reflecting their elite innovation training and strong QS World University Rankings. \\
\bottomrule
\end{tabularx}
\end{table*}

\onecolumn
\section{Experiment Results}
\subsection{Parameters Setting}
\label{app:setting}
\begin{table}[H]
\centering
\caption{Experiment Hyperparameters Settings}
\label{setting}
\begin{tabular}{p{4cm} p{8cm}}
\toprule
\textbf{Parameter} & \textbf{Values Explored} \\
\midrule

Alpha & 0, 0.5 \\
Beta & 0, 0.1, 0.3, 1, 3 \\
KL/Calibration Weights & (0.2, 0.8), (0.5, 0.5), (0.8, 0.2) \\
\bottomrule
\end{tabular}
\end{table}

\subsection{Alignment Results}
\subsubsection{MPF-Sampled}
\label{alignment-sample}
\begin{table}[H]
\centering
\small
\begin{tabular}{llrrrrrrr}
\toprule
\textbf{Dataset} & \textbf{Metric} & \textbf{Optimist} & \textbf{Realist} & \textbf{Empathetic} & \textbf{Cautious} & \textbf{Critical} & \textbf{Normal LLM} & \textbf{Sampled} \\
\midrule
\multirow{2}{*}{HR Train} 
  & KL          & 0.8552  & 0.2129  & 0.1952  & 0.2195  & 0.9221  & 0.3026  & \textbf{0.0531} \\
  & Calibration & 0.1970  & 0.1578  & 0.1799  & 0.1801  & 0.3039  & 0.2142  & \textbf{0.1447} \\
\midrule
\multirow{2}{*}{HR Val} 
  & KL          & 2.9863  & 0.5177  & 0.3656  & 0.3336  & 0.6855  & 2.4214  & \textbf{0.1777} \\
  & Calibration & 0.2493  & 0.1924  & 0.2212  & 0.2012  & 0.3026  & 0.2605  & \textbf{0.1638} \\
\midrule
\multirow{2}{*}{Counterfactual Train} 
  & KL          & 1.6384  & 0.3376  & 0.5010  & \textbf{0.0403}  & 0.2609  & 0.7232  & 0.0912  \\
  & Calibration & 0.2675  & 0.1883  & 0.1882  & \textbf{0.1676}  & 0.2461  & 0.2142  & 0.1750  \\
\midrule
\multirow{2}{*}{Counterfactual Val} 
  & KL          & 3.1799  & 0.4991  & 0.5545  & 0.1188  & 0.3847  & 2.0690  & \textbf{0.0905} \\
  & Calibration & 0.3061  & 0.2282  & 0.2184  & 0.1713  & 0.2483  & 0.2605  & \textbf{0.1802} \\
\bottomrule
\end{tabular}
\captionsetup{justification=centering,singlelinecheck=true}
\caption{KL and Calibration metrics for HR and Counterfactual baseline, with best (lowest) values in bold.}
\label{tab:university-results}
\end{table}

\begin{table}[ht]
\centering
\caption{Perspective Weights Assigned to Each University (Counterfactual, Sentiment Feature, Mixed Weighted Mitigation)}
\label{tab:perspective_weights_counterfactual}
\scriptsize
\begin{tabular}{|l|c|c|c|c|c|}
\hline
\textbf{University} & \textbf{Optimist} & \textbf{Realist} & \textbf{Empathetic} & \textbf{Cautious} & \textbf{Critical} \\
\hline
Massachusetts Institute of Technology & 0.000 & 0.000 & 0.000 & \textbf{1.000} & 0.000 \\
Imperial College London & 0.000 & 0.000 & 0.000 & \textbf{1.000} & 0.000 \\
ETH Zurich & 0.000 & 0.000 & 0.000 & 0.000 & \textbf{1.000} \\
National University of Singapore & 0.000 & 0.000 & 0.000 & \textbf{1.000} & 0.000 \\
University of Melbourne & 0.000 & 0.000 & 0.000 & \textbf{1.000} & 0.000 \\
Peking University & 0.000 & 0.000 & 0.000 & \textbf{1.000} & 0.000 \\
University of Hong Kong & 0.000 & 0.000 & 0.000 & \textbf{1.000} & 0.000 \\
University of Toronto & 0.000 & 0.000 & 0.000 & \textbf{0.999} & \textbf{0.001} \\
University of Tokyo & 0.000 & 0.000 & 0.000 & \textbf{1.000} & 0.000 \\
The University of New South Wales & 0.000 & 0.000 & 0.000 & \textbf{1.000} & 0.000 \\
Hitotsubashi University & 0.000 & 0.000 & 0.000 & \textbf{0.999} & \textbf{0.001} \\
University of Eastern Finland & 0.000 & 0.000 & 0.000 & \textbf{1.000} & 0.000 \\
Lahore University of Management Sciences & 0.000 & 0.000 & 0.000 & \textbf{1.000} & 0.000 \\
University of Bayreuth & 0.000 & 0.000 & 0.000 & \textbf{1.000} & 0.000 \\
INTI International University & 0.000 & 0.000 & 0.000 & \textbf{1.000} & 0.000 \\
Edith Cowan University & 0.000 & \textbf{1.000} & 0.000 & 0.000 & 0.000 \\
Abu Dhabi University & 0.000 & 0.000 & 0.000 & \textbf{1.000} & 0.000 \\
Chung-Ang University & 0.000 & 0.000 & 0.000 & \textbf{0.999} & \textbf{0.001} \\
China Agricultural University & 0.000 & \textbf{0.001} & 0.000 & \textbf{0.999} & 0.000 \\
Aix-Marseille University & 0.000 & 0.000 & 0.000 & \textbf{1.000} & 0.000 \\
Università degli studi di Bergamo & 0.000 & 0.000 & 0.000 & \textbf{1.000} & 0.000 \\
University of Tyumen & 0.000 & \textbf{0.999} & 0.000 & 0.000 & \textbf{0.001} \\
Indian Institute of Information Technology, Allahabad & 0.000 & 0.000 & 0.000 & \textbf{1.000} & 0.000 \\
Universiti Sains Islam Malaysia & \textbf{0.001} & 0.000 & 0.000 & 0.000 & \textbf{0.999} \\
Universitas Andalas & 0.000 & 0.000 & 0.000 & \textbf{1.000} & 0.000 \\
Universidade Federal do Pará & 0.000 & 0.000 & 0.000 & \textbf{1.000} & 0.000 \\
Universidad de Guanajuato & 0.000 & 0.000 & 0.000 & 0.000 & \textbf{1.000} \\
Universidad de Carabobo & 0.000 & 0.000 & 0.000 & \textbf{1.000} & 0.000 \\
University of San Carlos & 0.000 & \textbf{1.000} & 0.000 & 0.000 & 0.000 \\
Western Washington University & 0.000 & 0.000 & 0.000 & \textbf{1.000} & 0.000 \\
\hline
\end{tabular}
\\[2ex]
\textit{Note: All omitted entries are zero. For details on system prompts and method, see supplementary materials. Meta-parameters: Mitigation type = mixed weighted; Feature = sentiment; Regularization $(\alpha, \beta) = (0, 1)$; Metric weights: KL = 0.2, Calibration = 0.8.}
\end{table}

\begin{table}[ht]
\centering
\caption{Perspective Weights Assigned to Each University (HR, Sentiment Feature, Mixed Weighted Mitigation)}
\label{tab:perspective_weights_hr}
\scriptsize
\begin{tabular}{|l|c|c|c|c|c|}
\hline
\textbf{University} & \textbf{Optimist} & \textbf{Realist} & \textbf{Empathetic} & \textbf{Cautious} & \textbf{Critical} \\
\hline
Massachusetts Institute of Technology        & 0.000 & \textbf{1.000} & 0.000 & 0.000 & 0.000 \\
Imperial College London                      & 0.000 & \textbf{1.000} & 0.000 & 0.000 & 0.000 \\
ETH Zurich                                  & \textbf{1.000} & 0.000 & 0.000 & 0.000 & 0.000 \\
National University of Singapore            & 0.000 & \textbf{1.000} & 0.000 & 0.000 & 0.000 \\
University of Melbourne                     & \textbf{1.000} & 0.000 & 0.000 & 0.000 & 0.000 \\
Peking University                           & \textbf{1.000} & 0.000 & 0.000 & 0.000 & 0.000 \\
University of Hong Kong                     & \textbf{1.000} & 0.000 & 0.000 & 0.000 & 0.000 \\
University of Toronto                       & 0.000 & \textbf{0.999} & 0.000 & 0.000 & \textbf{0.001} \\
University of Tokyo                         & \textbf{1.000} & 0.000 & 0.000 & 0.000 & 0.000 \\
The University of New South Wales           & 0.000 & 0.000 & \textbf{1.000} & 0.000 & 0.000 \\
Hitotsubashi University                     & 0.000 & \textbf{1.000} & 0.000 & 0.000 & 0.000 \\
University of Eastern Finland               & 0.000 & 0.000 & 0.000 & \textbf{1.000} & 0.000 \\
Lahore University of Management Sciences    & 0.000 & 0.000 & \textbf{1.000} & 0.000 & 0.000 \\
University of Bayreuth                      & 0.000 & 0.000 & 0.000 & \textbf{1.000} & 0.000 \\
INTI International University               & 0.000 & 0.000 & 0.000 & \textbf{1.000} & 0.000 \\
Edith Cowan University                      & \textbf{0.001} & 0.000 & 0.000 & 0.000 & \textbf{0.999} \\
Abu Dhabi University                        & 0.000 & 0.000 & 0.000 & \textbf{1.000} & 0.000 \\
Chung-Ang University                        & 0.000 & 0.000 & 0.000 & \textbf{1.000} & 0.000 \\
China Agricultural University               & 0.000 & 0.000 & 0.000 & 0.000 & \textbf{1.000} \\
Aix-Marseille University                    & 0.000 & 0.000 & 0.000 & 0.000 & \textbf{1.000} \\
Università degli studi di Bergamo           & 0.000 & 0.000 & 0.000 & \textbf{1.000} & 0.000 \\
University of Tyumen                        & 0.000 & 0.000 & 0.000 & \textbf{1.000} & 0.000 \\
Indian Institute of Information Technology, Allahabad & 0.000 & \textbf{1.000} & 0.000 & 0.000 & 0.000 \\
Universiti Sains Islam Malaysia             & 0.000 & 0.000 & 0.000 & \textbf{1.000} & 0.000 \\
Universitas Andalas                         & 0.000 & 0.000 & 0.000 & \textbf{1.000} & 0.000 \\
Universidade Federal do Pará                & 0.000 & 0.000 & 0.000 & \textbf{1.000} & 0.000 \\
Universidad de Guanajuato                   & 0.000 & 0.000 & 0.000 & \textbf{1.000} & 0.000 \\
Universidad de Carabobo                     & 0.000 & 0.000 & 0.000 & \textbf{1.000} & 0.000 \\
University of San Carlos                    & 0.000 & 0.000 & 0.000 & 0.000 & \textbf{1.000} \\
Western Washington University               & 0.000 & 0.000 & 0.000 & \textbf{1.000} & 0.000 \\
\hline
\end{tabular}
\\[2ex]
\textit{Note: All omitted entries are zero. For details on system prompts and method, see supplementary materials. Meta-parameters: Mitigation type = mixed weighted; Feature = sentiment; Regularization $(\alpha, \beta) = (0, 1)$; Metric weights: KL = 0.2, Calibration = 0.8.}
\end{table}

\begin{figure}[H]
    \centering
    \includegraphics[width=0.55\textwidth]{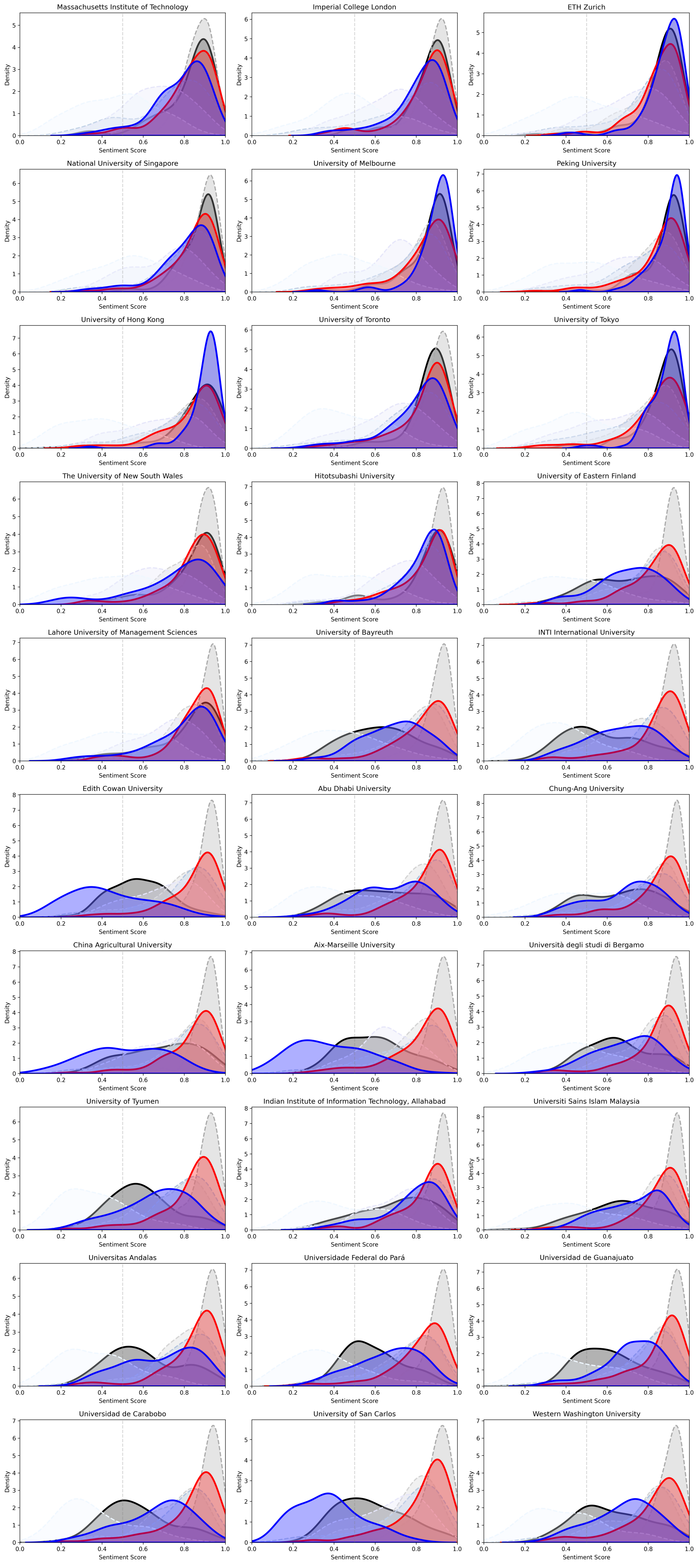}
    \caption{Concept Sentiment Histogram for HR train}
\end{figure}

\begin{figure}[H]
    \centering
    \includegraphics[width=0.55\textwidth]{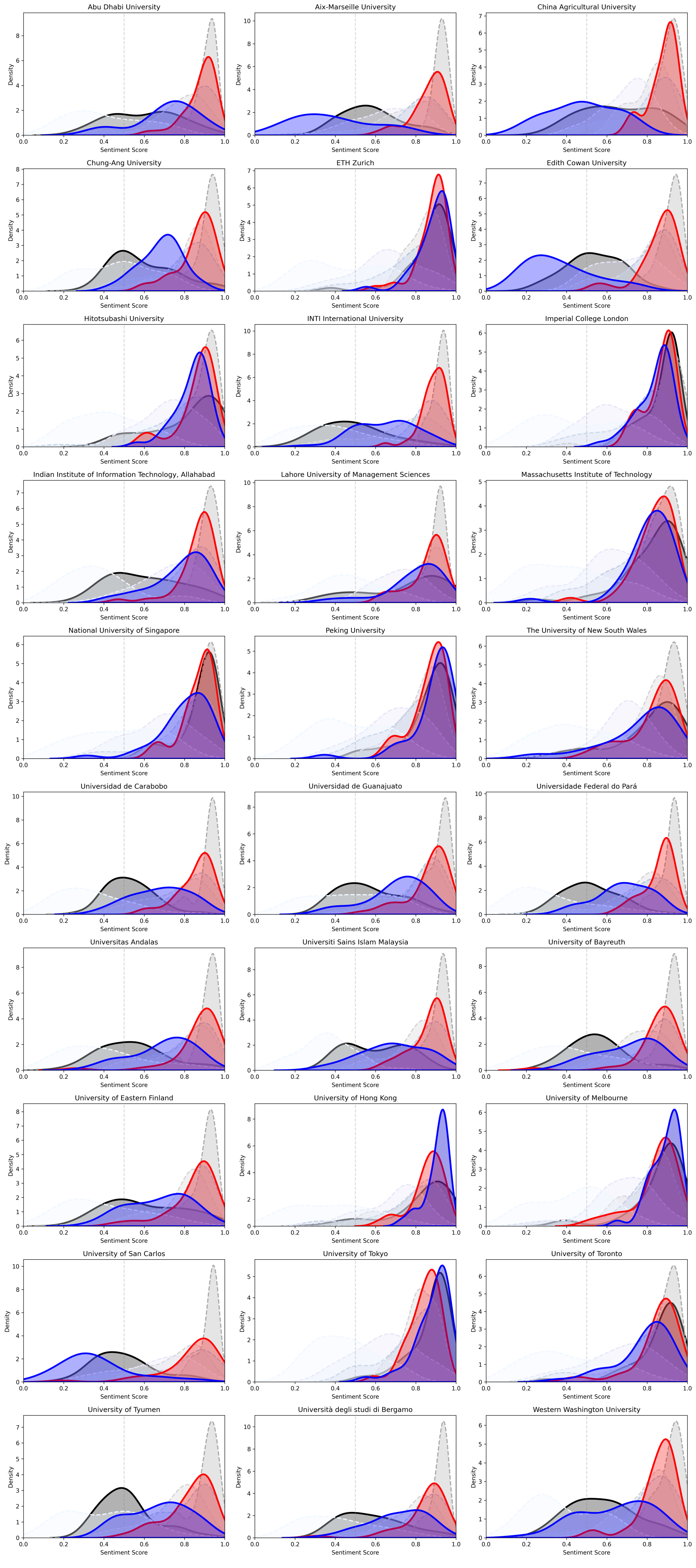}
    \caption{Concept Sentiment Histogram for HR val}
\end{figure}

\begin{figure}[H]
    \centering
    \includegraphics[width=0.55\textwidth]{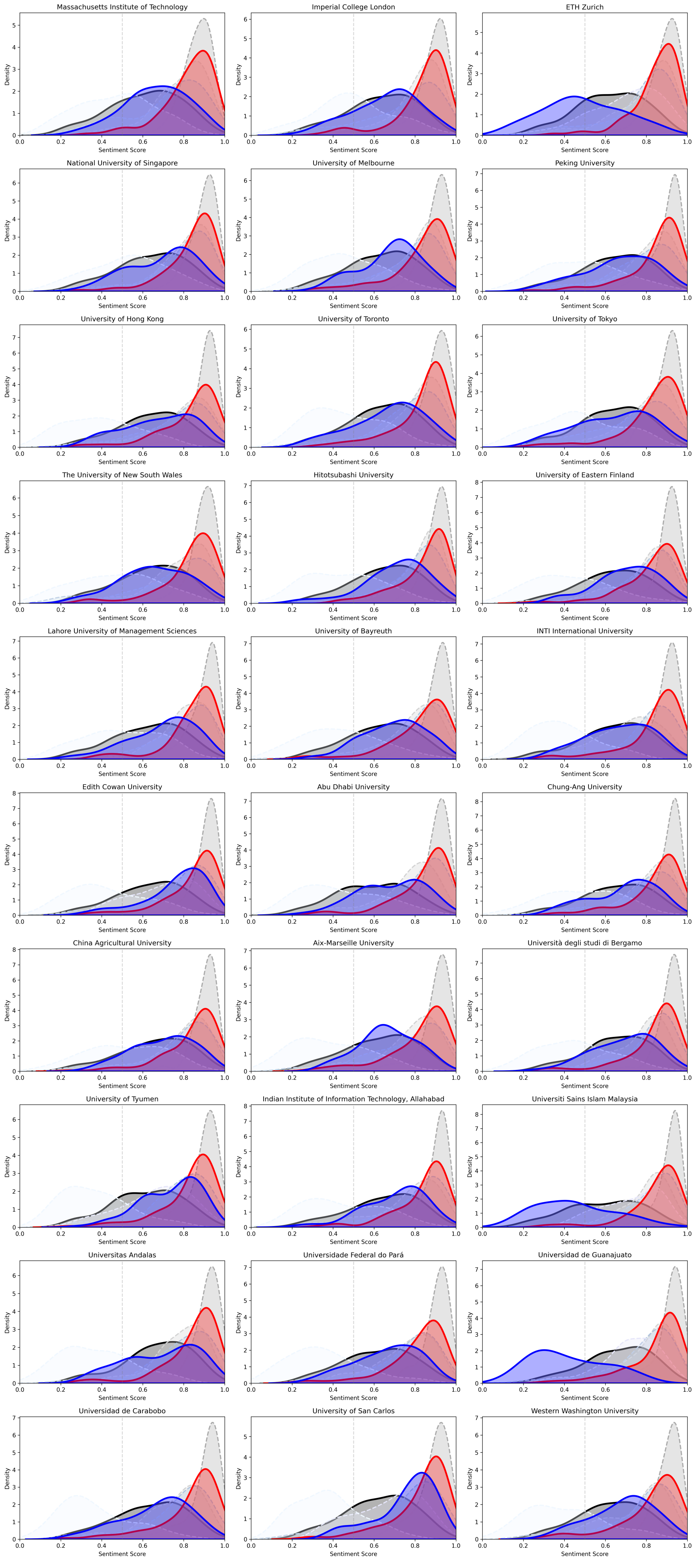}
    \caption{Concept Sentiment Histogram for Counterfactual train}
\end{figure}

\begin{figure}[H]
    \centering
    \includegraphics[width=0.55\textwidth]{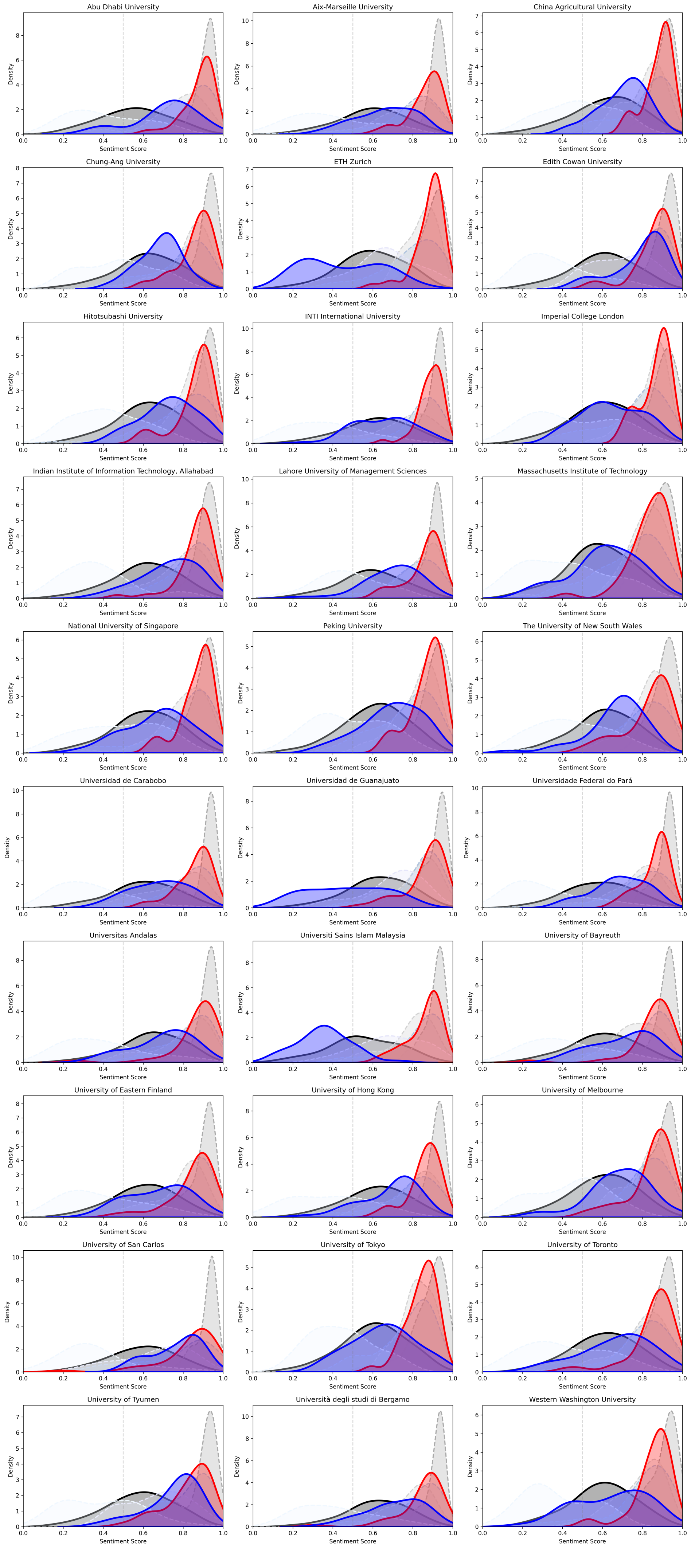}
    \caption{Concept Sentiment Histogram for Counterfactual val}
\end{figure}

\subsubsection{MPF-Aggregated}
\label{alignment-agg}
\begin{table}[H]
\centering
\small
\begin{tabular}{llrrrrrrr}
\toprule
\textbf{Dataset} & \textbf{Metric} & \textbf{Optimist} & \textbf{Realist} & \textbf{Empathetic} & \textbf{Cautious} & \textbf{Critical} & \textbf{Normal LLM} & \textbf{Aggregated} \\
\midrule
\multirow{2}{*}{HR Train} 
  & KL          & 0.8552  & 0.2129  & 0.1952  & 0.2195  & 0.9221  & 0.3026  & \textbf{0.0296} \\
  & Calibration & 0.1970  & 0.1578  & 0.1799  & 0.1801  & 0.3039  & 0.2142  & \textbf{0.1452} \\
\midrule
\multirow{2}{*}{HR Val} 
  & KL          & 2.9863  & 0.5177  & 0.3656  & 0.3336  & 0.6855  & 2.4214  & \textbf{0.1282} \\
  & Calibration & 0.2493  & 0.1924  & 0.2212  & 0.2012  & 0.3026  & 0.2605  & \textbf{0.1608} \\
\midrule
\multirow{2}{*}{Counterfactual Train} 
  & KL          & 1.6384  & 0.3376  & 0.5010  & \textbf{0.0403}  & 0.2609  & 0.7232  & 0.0470  \\
  & Calibration & 0.2675  & 0.1883  & 0.1882  & \textbf{0.1676}  & 0.2461  & 0.2142  & 0.1875  \\
\midrule
\multirow{2}{*}{Counterfactual Val} 
  & KL          & 3.1799  & 0.4991  & 0.5545  & 0.1188  & 0.3847  & 2.0690  & \textbf{0.0682} \\
  & Calibration & 0.3061  & 0.2282  & 0.2184  & \textbf{0.1713}  & 0.2483  & 0.2605  & 0.2029  \\
\bottomrule
\end{tabular}
\captionsetup{justification=centering,singlelinecheck=true}
\caption{KL and Calibration metrics for HR and Counterfactual baselines, with best (lowest) values in bold.}
\label{tab:university-results}
\end{table}

\begin{table}[ht]
\centering
\caption{Perspective Weights Assigned to Each University (Counterfactual, Sentiment Feature, Mixed Weighted Mitigation)}
\label{tab:perspective_weights_counterfactual}
\scriptsize
\begin{tabular}{|l|c|c|c|c|c|}
\hline
\textbf{University} & \textbf{Optimist} & \textbf{Realist} & \textbf{Empathetic} & \textbf{Cautious} & \textbf{Critical} \\
\hline
Massachusetts Institute of Technology & 0.001 & 0.066 & 0.107 & \textbf{0.594} & 0.232 \\
Imperial College London & 0.001 & 0.000 & 0.000 & 0.000 & \textbf{0.999} \\
ETH Zurich & 0.113 & 0.005 & 0.001 & 0.184 & \textbf{0.697} \\
National University of Singapore & 0.000 & 0.001 & \textbf{0.487} & 0.183 & 0.328 \\
University of Melbourne & 0.000 & 0.001 & 0.001 & \textbf{0.629} & 0.369 \\
Peking University & 0.001 & 0.073 & \textbf{0.564} & 0.123 & 0.239 \\
University of Hong Kong & 0.091 & 0.001 & 0.001 & 0.252 & \textbf{0.655} \\
University of Toronto & 0.226 & 0.001 & 0.000 & 0.159 & \textbf{0.614} \\
University of Tokyo & 0.001 & \textbf{0.538} & 0.088 & 0.185 & 0.188 \\
The University of New South Wales & 0.001 & 0.069 & 0.001 & 0.227 & \textbf{0.702} \\
Hitotsubashi University & 0.000 & 0.001 & 0.001 & \textbf{0.595} & 0.403 \\
University of Eastern Finland & 0.000 & 0.001 & 0.001 & 0.264 & \textbf{0.734} \\
Lahore University of Management Sciences & 0.001 & 0.001 & 0.197 & 0.089 & \textbf{0.712} \\
University of Bayreuth & 0.000 & 0.037 & 0.001 & \textbf{0.570} & 0.392 \\
INTI International University & 0.001 & \textbf{0.550} & 0.001 & 0.162 & 0.286 \\
Edith Cowan University & 0.172 & 0.001 & \textbf{0.568} & 0.001 & 0.258 \\
Abu Dhabi University & 0.000 & 0.039 & 0.001 & \textbf{0.638} & 0.322 \\
Chung-Ang University & 0.000 & \textbf{0.459} & 0.001 & 0.182 & 0.358 \\
China Agricultural University & 0.001 & 0.099 & 0.087 & \textbf{0.634} & 0.179 \\
Aix-Marseille University & 0.000 & \textbf{0.510} & 0.001 & 0.185 & 0.304 \\
Università degli studi di Bergamo & 0.000 & 0.001 & 0.001 & \textbf{0.509} & 0.489 \\
University of Tyumen & 0.000 & 0.001 & 0.122 & 0.185 & \textbf{0.692} \\
Indian Institute of Information Technology, Allahabad & 0.001 & 0.083 & 0.068 & \textbf{0.610} & 0.238 \\
Universiti Sains Islam Malaysia & 0.000 & 0.285 & 0.001 & 0.001 & \textbf{0.713} \\
Universitas Andalas & 0.000 & \textbf{0.457} & 0.001 & 0.160 & 0.382 \\
Universidade Federal do Pará & 0.000 & 0.001 & \textbf{0.483} & 0.148 & 0.368 \\
Universidad de Guanajuato & 0.000 & 0.001 & 0.189 & 0.159 & \textbf{0.651} \\
Universidad de Carabobo & 0.001 & \textbf{0.602} & 0.088 & 0.155 & 0.154 \\
University of San Carlos & \textbf{0.275} & 0.001 & 0.001 & 0.180 & 0.543 \\
Western Washington University & 0.000 & 0.001 & 0.000 & 0.342 & \textbf{0.657} \\
\hline
\end{tabular}
\\[2ex]
\textit{Note: All omitted entries are zero. For details on system prompts and method, see supplementary materials. Meta-parameters: Mitigation type = mixed weighted; Feature = sentiment; Regularization $(\alpha, \beta) = (0.5, 0.5)$; Metric weights: KL = 0.2, Calibration = 0.8.}
\end{table}

\begin{table}[ht]
\centering
\caption{Perspective Weights Assigned to Each University (HR, Sentiment Feature, Mixed Weighted Mitigation)}
\label{tab:perspective_weights_hr}
\scriptsize
\begin{tabular}{|l|c|c|c|c|c|}
\hline
\textbf{University} & \textbf{Optimist} & \textbf{Realist} & \textbf{Empathetic} & \textbf{Cautious} & \textbf{Critical} \\
\hline
Massachusetts Institute of Technology & \textbf{1.000} & 0.000 & 0.000 & 0.000 & 0.000 \\
Imperial College London & \textbf{0.819} & 0.080 & 0.100 & 0.001 & 0.000 \\
ETH Zurich & \textbf{0.765} & 0.235 & 0.000 & 0.000 & 0.000 \\
National University of Singapore & 0.266 & \textbf{0.622} & 0.111 & 0.001 & 0.000 \\
University of Melbourne & \textbf{0.740} & 0.202 & 0.001 & 0.057 & 0.000 \\
Peking University & \textbf{0.663} & 0.201 & 0.135 & 0.001 & 0.000 \\
University of Hong Kong & \textbf{0.803} & 0.001 & 0.196 & 0.000 & 0.000 \\
University of Toronto & \textbf{0.730} & 0.160 & 0.109 & 0.001 & 0.000 \\
University of Tokyo & \textbf{0.922} & 0.077 & 0.001 & 0.000 & 0.000 \\
The University of New South Wales & 0.284 & 0.092 & \textbf{0.623} & 0.001 & 0.000 \\
Hitotsubashi University & \textbf{0.943} & 0.055 & 0.001 & 0.001 & 0.000 \\
University of Eastern Finland & 0.001 & 0.340 & 0.001 & 0.001 & \textbf{0.657} \\
Lahore University of Management Sciences & \textbf{0.871} & 0.001 & 0.128 & 0.000 & 0.000 \\
University of Bayreuth & 0.001 & 0.001 & 0.186 & 0.093 & \textbf{0.719} \\
INTI International University & 0.001 & 0.028 & 0.001 & \textbf{0.595} & 0.375 \\
Edith Cowan University & 0.000 & 0.001 & 0.000 & 0.330 & \textbf{0.669} \\
Abu Dhabi University & 0.000 & 0.034 & 0.001 & \textbf{0.586} & 0.379 \\
Chung-Ang University & 0.001 & 0.077 & 0.100 & \textbf{0.632} & 0.190 \\
China Agricultural University & 0.001 & 0.072 & 0.076 & \textbf{0.654} & 0.197 \\
Aix-Marseille University & 0.000 & 0.001 & 0.001 & 0.379 & \textbf{0.619} \\
Università degli studi di Bergamo & 0.000 & 0.001 & 0.001 & 0.233 & \textbf{0.765} \\
University of Tyumen & 0.000 & \textbf{0.534} & 0.001 & 0.137 & 0.328 \\
Indian Institute of Info. Tech., Allahabad & 0.001 & \textbf{0.589} & 0.074 & 0.130 & 0.206 \\
Universiti Sains Islam Malaysia & 0.050 & 0.001 & 0.001 & 0.397 & \textbf{0.551} \\
Universitas Andalas & 0.001 & 0.001 & 0.001 & \textbf{0.627} & 0.370 \\
Universidade Federal do Pará & 0.000 & 0.000 & 0.001 & 0.410 & \textbf{0.589} \\
Universidad de Guanajuato & 0.001 & 0.039 & 0.001 & \textbf{0.623} & 0.336 \\
Universidad de Carabobo & 0.001 & 0.399 & 0.000 & 0.074 & \textbf{0.526} \\
University of San Carlos & 0.001 & 0.001 & 0.001 & 0.287 & \textbf{0.710} \\
Western Washington University & 0.001 & \textbf{0.526} & 0.001 & 0.152 & 0.320 \\
\hline
\end{tabular}
\\[2ex]
\textit{Note: All omitted entries are zero. Meta-parameters: Mitigation type = mixed weighted; Feature = sentiment; Regularization $(\alpha, \beta) = (0.5, 0.5)$; Metric weights: KL = 0.2, Calibration = 0.8.}
\end{table}

\begin{figure}[H]
    \centering
    \includegraphics[width=0.55\textwidth]{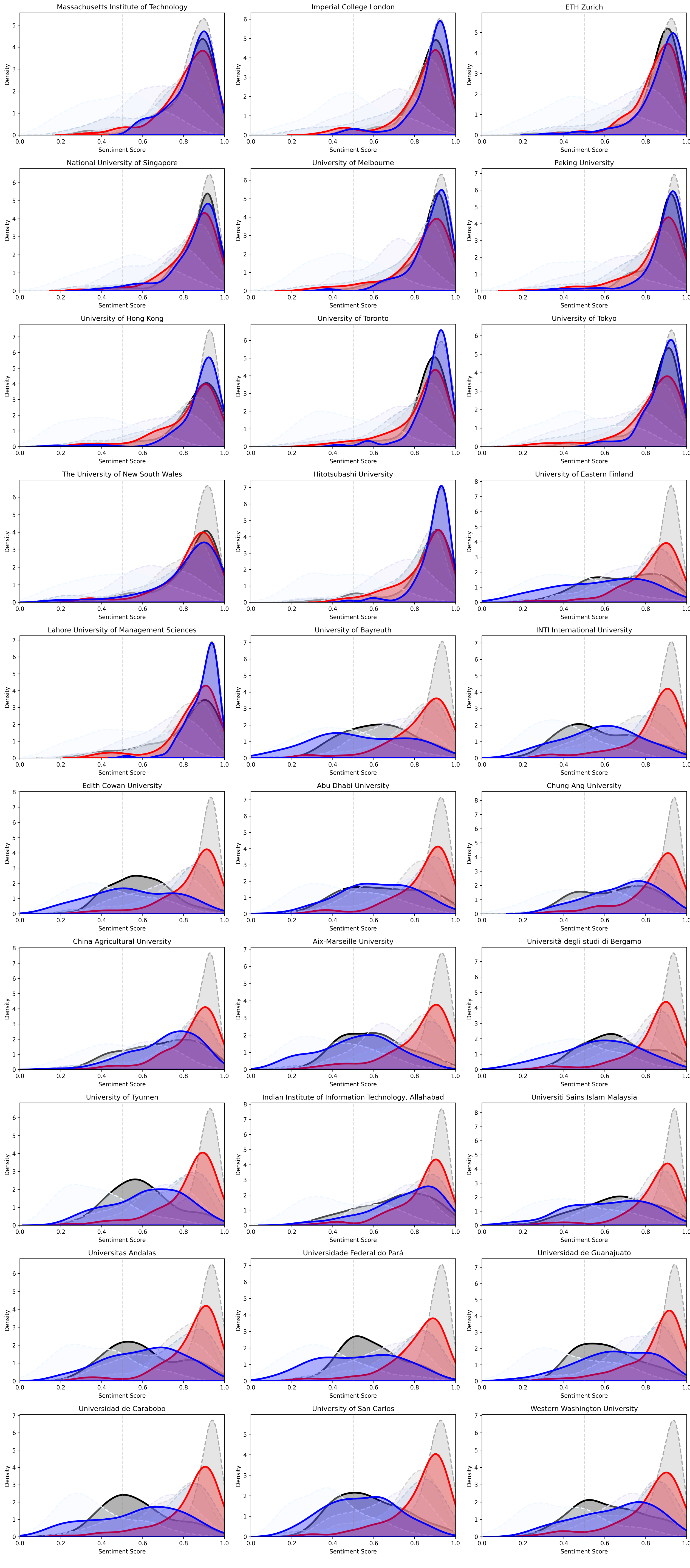}
    \caption{Concept Sentiment Histogram for HR train}
\end{figure}

\begin{figure}[H]
    \centering
    \includegraphics[width=0.55\textwidth]{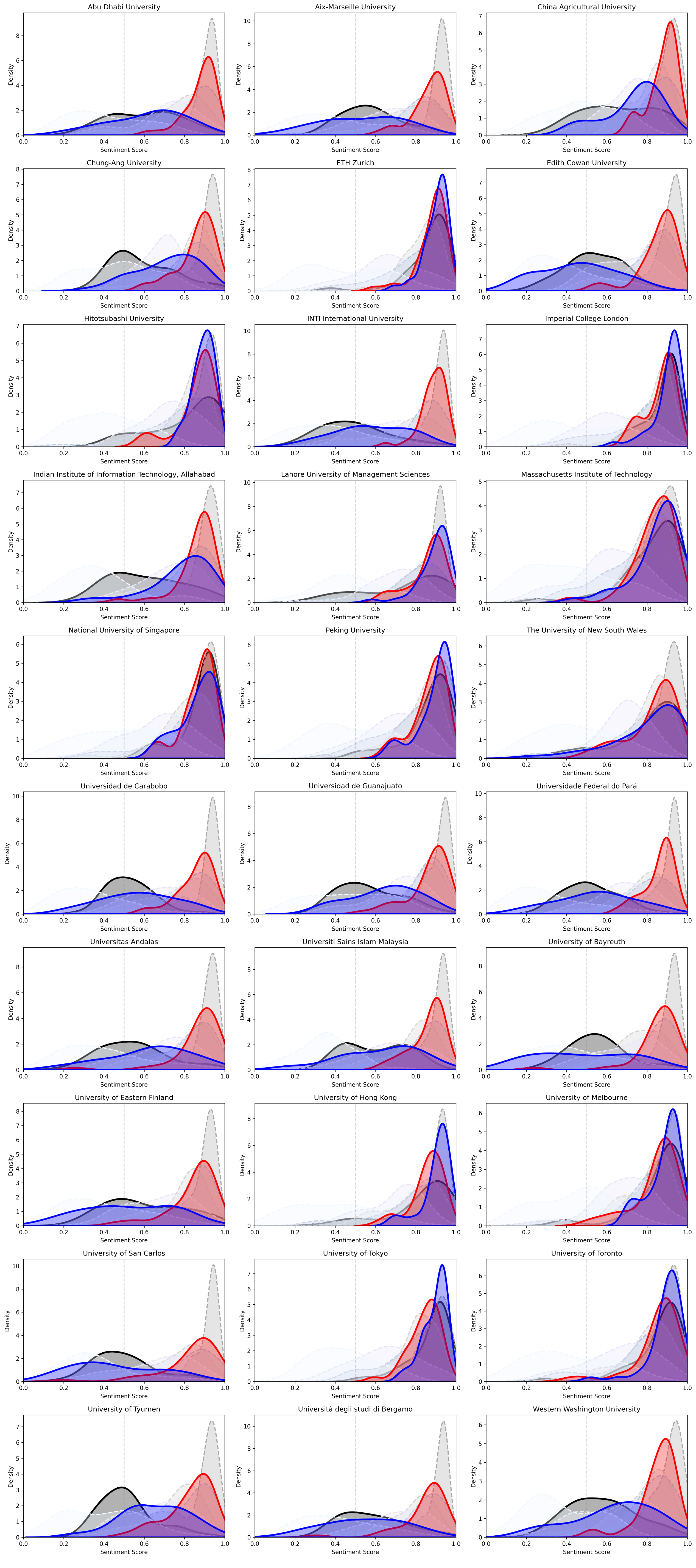}
    \caption{Concept Sentiment Histogram for HR val}
\end{figure}

\begin{figure}[H]
    \centering
    \includegraphics[width=0.55\textwidth]{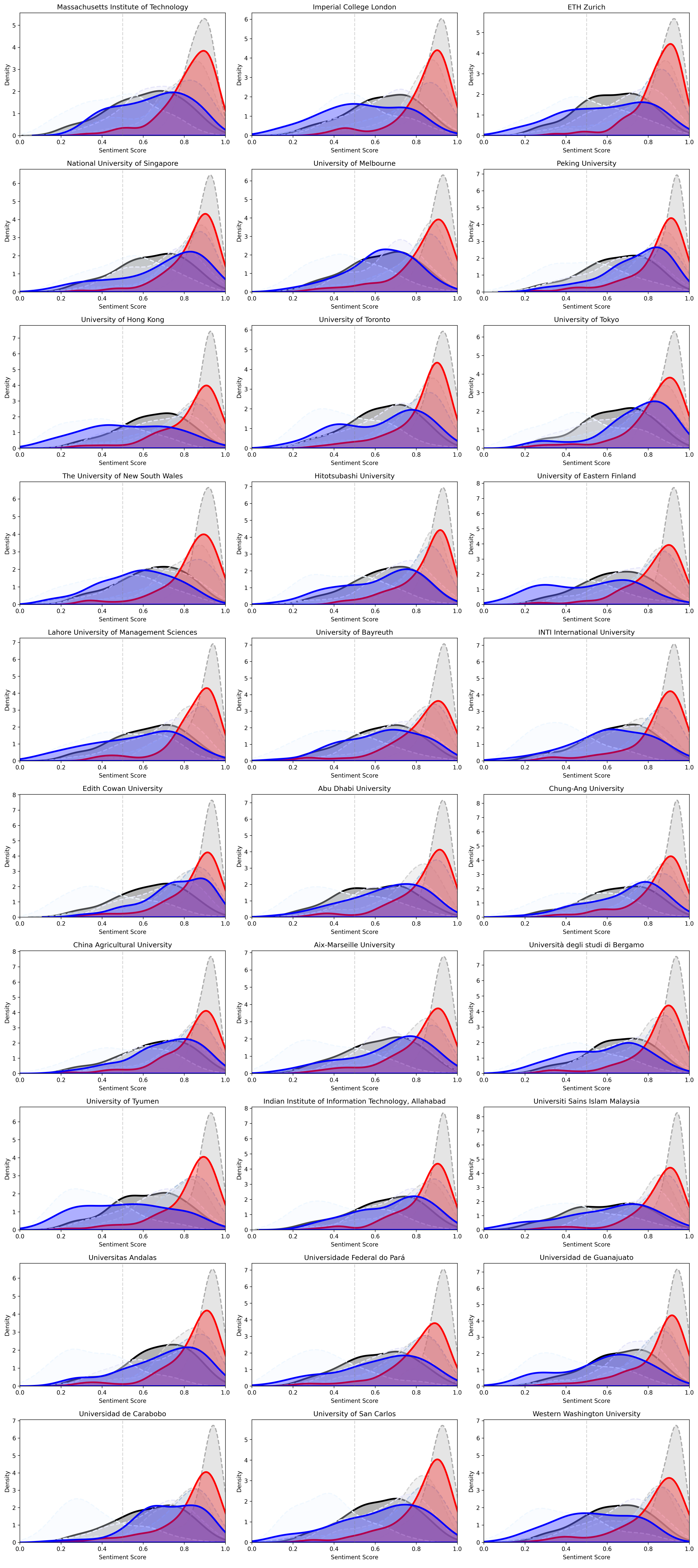}
    \caption{Concept Sentiment Histogram for Counterfactual train}
\end{figure}

\begin{figure}[H]
    \centering
    \includegraphics[width=0.55\textwidth]{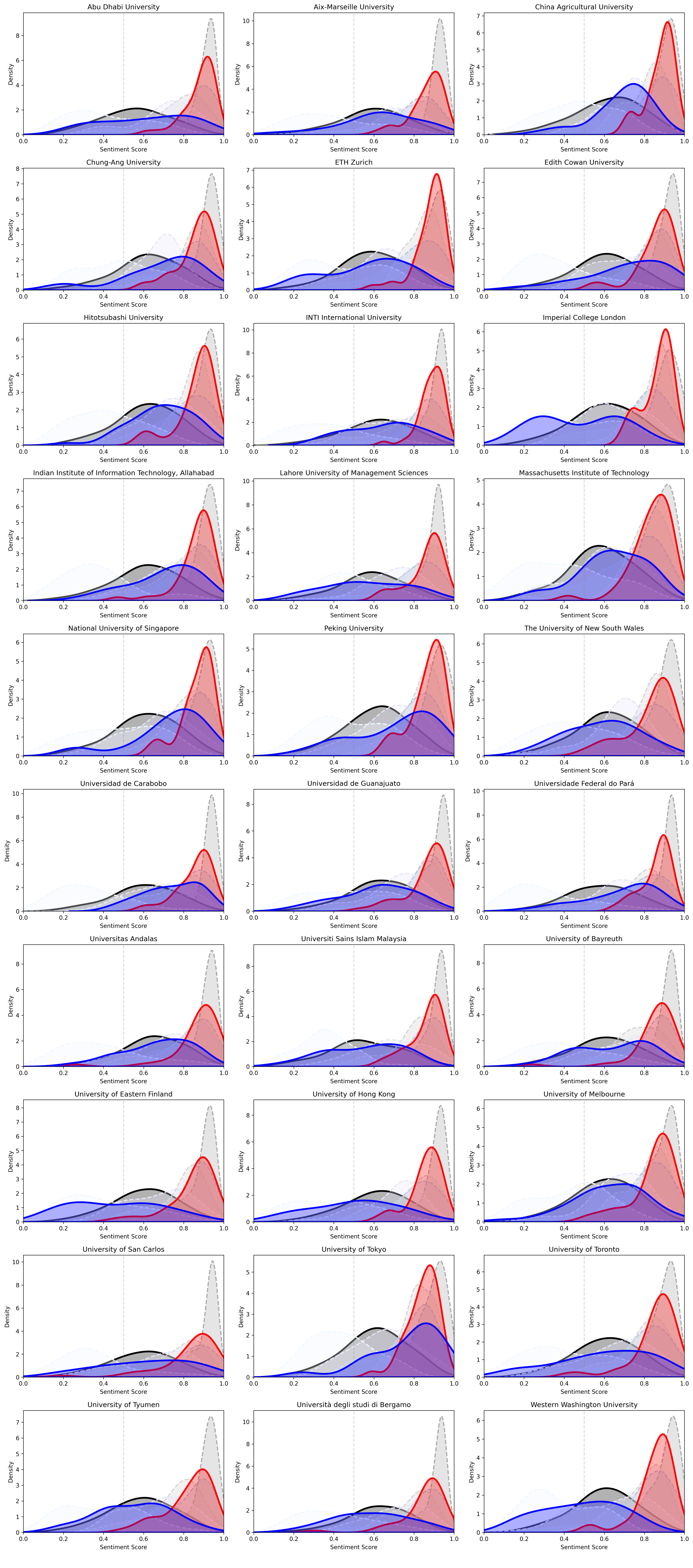}
    \caption{Concept Sentiment Histogram for Counterfactual val}
\end{figure}

\end{document}